\documentclass[journal]{IEEEtran}

\usepackage[hyphens]{url}
\usepackage{graphicx}
\usepackage{amsmath,amssymb}
\usepackage{booktabs}
\usepackage{multirow}
\usepackage{tabularx}
\usepackage{array}
\usepackage{cite}
\usepackage[hidelinks]{hyperref}

\usepackage{caption}

\newcommand{\authorbox}[3]{%
    \begin{minipage}[t]{0.235\textwidth}
        \centering
        \normalsize #1\par
        \vspace{0.15em}
        {\footnotesize\itshape #2\par}
        {\footnotesize #3\par}
    \end{minipage}%
}

\title{Uncovering and Mitigating Positional Blind Spots in\\
Vision-Language-Action Models}

\author{%
\makebox[\textwidth][c]{%
\begin{minipage}{\textwidth}
    \centering

    \noindent
    \authorbox{Dongdong An}{Shanghai Normal University}{Shanghai, China}
    \hfill
    \authorbox{Pengjie Zhao}{Shanghai Normal University}{Shanghai, China}
    \hfill
    \authorbox{Yihao Huang}{East China Normal University}{Shanghai, China}
    \hfill
    \authorbox{Wenbing Tang}{Northwest A\&F University}{Yangling, Shaanxi, China}

    \par\vspace{1.1em}

    \noindent
    \authorbox{Ziming He}{Northwest A\&F University}{Yangling, Shaanxi, China}
    \hfill
    \authorbox{Jiayi Zhu}{Xidian University}{Xi'an, Shaanxi, China}
    \hfill
    \authorbox{Jifeng Ning}{Northwest A\&F University}{Yangling, Shaanxi, China}
    \hfill
    \authorbox{Qin Zhao}{Shanghai Normal University}{Shanghai, China}

\end{minipage}%
}}

\begin{document}

\maketitle

\begin{abstract}
Recent Vision-Language-Action (VLA) models achieve promising performance in robotic manipulation, typically measured by success rates aggregated over predefined object configurations, an evaluation that implicitly assumes spatially uniform competence across the workspace.
However, this assumption does not hold: even with the instruction and every other scene factor held fixed, merely relocating a task-irrelevant distractor can sharply raise the failure probability within localized, spatially coherent regions, which we term \emph{Positional Blind Spots} (PBS).
In this paper, we propose a two-stage black-box framework to uncover and mitigate PBS. During the uncovering stage, we grid the workspace and apply a one-sided log-likelihood-ratio test to localize PBS cells with significantly elevated risk. During the mitigation stage, we fine-tune the policy via LoRA on demonstrations collected from these PBS regions, improving competence there while largely preserving performance across the rest of the workspace.
We evaluate our framework on five state-of-the-art VLA policies across two benchmarks, and find that PBS are pervasive and spatially concentrated in all of them, with failure rates up to 0.58. Our search strategy achieves an average F1-score of 0.678, outperforming random search and adaptive sampling baselines by 0.268 and 0.178, respectively. Guided by the discovered regions, targeted fine-tuning reduces the overall failure rate by 40.00\%--85.19\%.

\end{abstract}

\section{Introduction}

Enabling robots to follow natural language instructions from raw visual input has long been a central goal of robotic manipulation, and Vision-Language-Action (VLA) models have emerged as the leading end-to-end approach toward this goal by mapping visual observations and language instructions directly to executable control actions~\cite{ma2026survey,li2026towards}.
Built upon pre-trained foundation models and fine-tuned on large-scale robot demonstrations, VLA models integrate perception, reasoning, and action within a unified architecture, and have reported impressive success rates across a wide range of manipulation benchmarks~\cite{zhang2025vla,xu2026vla}. 
Beyond manipulation, the VLA paradigm is rapidly extending to broader embodied domains, such as autonomous driving~\cite{yang2026drivemoe,jiang2025survey} and aerial navigation~\cite{chen2026aerialvla}. 
In these scenarios, the failure cost is no longer a benchmark statistic: an unexpected error may interrupt the task, damage objects, or even endanger humans. Therefore, reliability has become a prerequisite, rather than an afterthought, for bringing VLA models into the physical world.

\begin{figure}[t]
    \centering
    \includegraphics[
        width=\columnwidth,
    ]{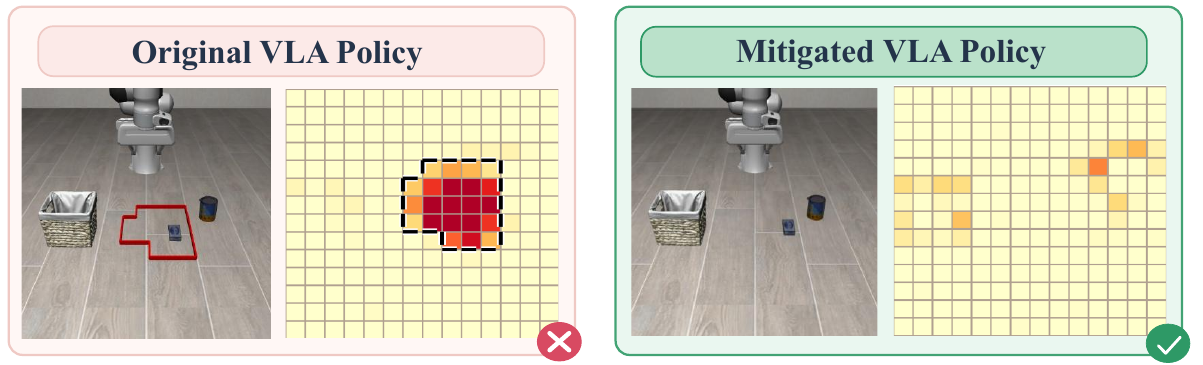}
    \caption{Illustration of Positional Blind Spots (PBS) in VLA policies and PBS-guided mitigation. The heatmaps visualize failure rates conditioned on distractor positions, revealing localized positional vulnerabilities.}
    \label{fig:intro}
\end{figure}

To assess the reliability of VLA models, existing evaluations mainly rely on predefined benchmarks that report average success rates over fixed tasks and environments~\cite{liu2023libero,zhang2025vla,fan2026safevla}. Recent studies further investigate model robustness under various input variations, such as changes in object appearance, language instructions, or scene configurations~\cite{wang2025vlatest,valle2026metamorphic,wang2025exploring,yan2025alignment}. However, these evaluations largely overlook a fundamental factor in embodied tasks: object position.
Since deployed objects can appear anywhere in the workspace rather than at the positions covered by training or evaluation, this gap raises a fundamental question left unanswered: does a VLA model's competence hold uniformly across the workspace, or collapse in specific regions, a phenomenon we term \emph{Positional Blind Spots (PBS)}.

It raises a question that can PBS be systematically discovered. Identifying such spatial failures is challenging due to the continuous nature and large search space of the robot workspace. Unlike conventional testing where the input space can often be enumerated or discretized, the possible object positions in embodied tasks form a continuous space with infinitely many possible configurations. Meanwhile, PBS may only occupy a small and localized portion of the workspace, making them difficult to reveal through sparse or uniform sampling. Moreover, evaluating a VLA policy typically requires executing complete task rollouts and only provides binary success or failure feedback, without gradients, confidence scores, or internal representations to guide the search. Consequently, exhaustive evaluation over fine-grained spatial regions is prohibitively expensive, while limited evaluations may easily miss critical blind spots. This raises a fundamental challenge: how to efficiently explore the continuous workspace under a limited evaluation budget and reliably identify high-risk regions where VLA models exhibit positional failures.

In this work, we propose a two-stage framework that uncovers and mitigates PBS in VLA policies without accessing their internals. During the uncovering stage, we address the challenge of searching over a continuous spatial space by discretizing the feasible object-placement workspace into a finite set of spatial regions. Specifically, we partition the workspace into a spatial grid, sample rollout points uniformly within each cell, and apply a Log-Likelihood Ratio (LLR) test to identify cells whose failure rates are significantly higher than the overall workspace distribution. These cells are returned as the predicted blind-spot regions. During the mitigation stage, human experts teleoperate the robot arm to collect additional successful demonstrations within these identified regions. The policy is then fine-tuned via LoRA following its original adaptation procedure, without modifying its observation, action, or optimization interface. In this way, the framework uses PBS-guided demonstrations to improve spatial reliability while preserving the original policy interface.

We evaluate PBS characterization, search, and mitigation on five representative VLA policies ($\pi_0$, $\pi_{0.5}$, OpenVLA-OFT, UniVLA, and VLA-Adapter) across two benchmarks (LIBERO and VLA-Arena).
Exhaustive spatial evaluation confirms that PBS are pervasive: after introducing a single task-irrelevant distractor, every evaluated policy exhibits a failure rate ranging from 0.15 to 0.58, with failures spatially concentrated rather than uniformly distributed across the workspace.
Under the same search budget, our search strategy recovers these regions substantially more accurately than random and adaptive sampling baselines, improving the average F1-score by 0.268 and 0.178, respectively.

In summary, our main contributions are as follows:
\begin{itemize}
    \item We identify and formalize \emph{Positional Blind Spots} (PBS), a previously unexplored spatial vulnerability of VLA policies where the presence of task-irrelevant objects at specific workspace positions induces systematic failures.
    \item We propose a two-stage black-box framework that uncovers PBS through spatial discretization and log-likelihood-ratio filtering, and mitigates them through search-guided policy fine-tuning, without modifying the policy architecture or interfaces.
    \item Through exhaustive spatial evaluations on five VLA policies across two benchmarks, we show that PBS consistently emerge across the evaluated settings, with failure rates up to 58\%, and that targeted mitigation reduces the overall failure rate by up to 85\%.
\end{itemize}

\section{Related Work}

\paragraph{Vision-Language-Action (VLA) models.} VLA models map visual observations and language instructions to robot actions by adapting pretrained vision-language representations with robot demonstrations. 
RT-2~\cite{zitkovich2023rt} casts actions as language-like tokens, while OpenVLA~\cite{kim2024openvla} scales this paradigm with diverse robot data and efficient downstream adaptation. 
Recent models adopt diverse action-generation and adaptation mechanisms: $\pi_0$~\cite{BlackK-2025pi_0} uses a flow-matching action expert, $\pi_{0.5}$~\cite{intelligence2025pi05} introduces heterogeneous co-training, OpenVLA-OFT~\cite{KimM1-oft-2025} combines continuous actions with action chunking and parallel decoding, UniVLA~\cite{BuQ-univla-25} learns transferable task-centric representations, and VLA-Adapter~\cite{wang_vla-adapter_2026} connects vision-language features to action prediction through a lightweight adapter. 
Broadly, these models cover distinct design directions in action generation and policy adaptation. $\pi_0$ and OpenVLA-OFT emphasize continuous or chunked action generation, while $\pi_{0.5}$, UniVLA, and VLA-Adapter explore heterogeneous co-training, transferable task representations, and lightweight adaptation, respectively.
Despite rapid progress in model architectures and adaptation strategies, ensuring robust and reliable behavior of VLA policies under diverse real-world variations remains an open challenge. In real-world manipulation scenarios, objects and environmental conditions may vary continuously, requiring VLA policies to maintain consistent performance beyond the configurations observed during training.

\paragraph{Evaluation and enhancement of VLA models.} Benchmarks such as LIBERO~\cite{liu2023libero}, VLABench~\cite{Zhang_2025_vlabench}, and VLA-Arena~\cite{zhang2026vlaarena} assess transfer, semantic and long-horizon manipulation, and robustness under structured task, language, and visual variations. 
Beyond benchmark-based evaluation, active testing methods further expose failures through scene mutations or semantics-preserving transformations~\cite{wang2025vlatest,valle2026metamorphic}, while adversarial studies employ textual, visual, and cross-modal perturbations~\cite{jones2025adversarial,wang2025exploring,yan2025alignment}. 
However, these approaches either evaluate performance over predefined task distributions, modify multiple factors simultaneously, or introduce intentionally crafted perturbations; they do not explicitly isolate the effect of task-irrelevant object relocation on VLA policy competence.
Existing enhancement methods rely on additional demonstrations or task adaptation, while safety-oriented studies emphasize alignment~\cite{zhang2025safevla} and failure interception~\cite{nips2025safe}. 
Robustness-oriented work has only recently considered physical sensor attacks~\cite{lu2026phantom}, leaving natural, task-preserving positional variations in manipulation largely unexplored.
However, these methods do not directly connect localized positional diagnosis with targeted data collection.
Different from existing approaches, our work investigates spatially concentrated positional vulnerabilities through black-box rollouts and uses the discovered regions to guide targeted demonstration collection and policy refinement.

\section{Preliminaries}

\subsection{Motivation}
\begin{figure}[t]
    \centering
    \includegraphics[
        width=\columnwidth,
    ]{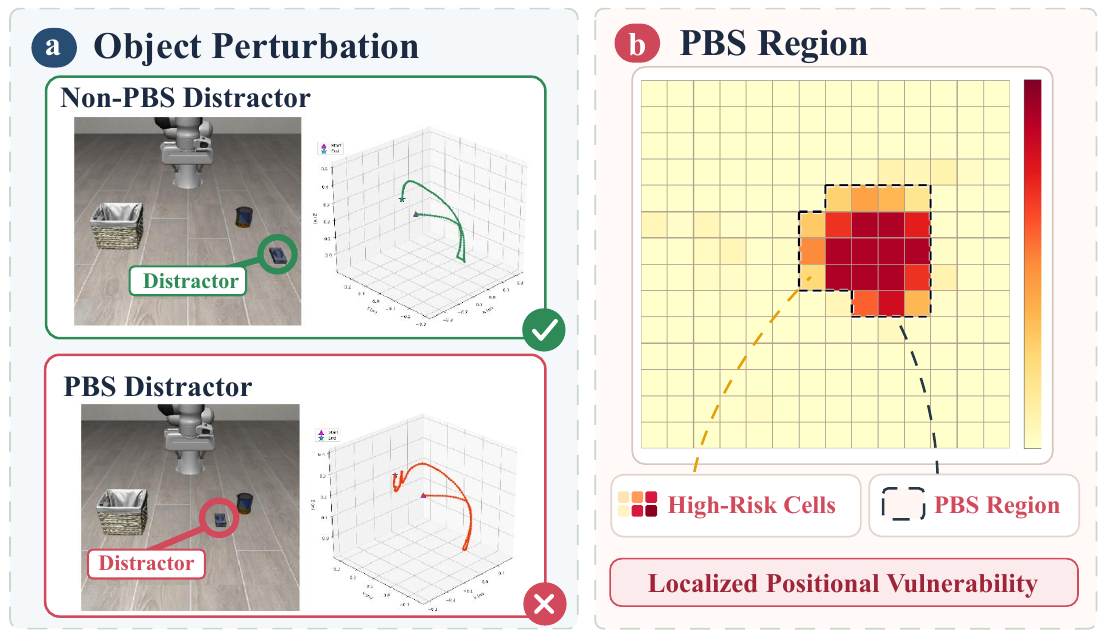}
    \caption{Impact of distractor position on VLA task success.}
    \label{fig:motivation}
\end{figure}

We first examine the positional sensitivity of VLA policies through a concrete example.
As shown in Figure~\ref{fig:motivation}(a), the policy successfully completes the task when a task-irrelevant distractor is placed at one feasible position.
However, relocating the same distractor to another feasible position causes the policy to fail.
Across the two executions, the task objective, language instruction, task-relevant object configuration, robot initialization, camera view, and distractor attributes remain unchanged.
Therefore, the presence of the distractor alone cannot explain the failure; instead, the behavioral difference is associated with the spatial position of the distractor in the workspace.


We further examine distractor placements across the feasible workspace to determine whether this positional sensitivity occurs only at isolated coordinates or forms spatially structured regions.
Figure~\ref{fig:motivation}(b) shows that positions where the distractor induces elevated failure rates form a localized and spatially coherent region, with neighboring placements exhibiting similar failure risks.
A policy may therefore remain reliable when the distractor is placed in most regions of the workspace while exhibiting substantially higher failure probability when the same distractor is placed within a small subset of positions. Such localized vulnerabilities can be obscured by aggregate task-level success rates, which average performance across low- and high-risk locations, despite their practical relevance when task-irrelevant objects appear at diverse feasible positions during deployment.


We term these localized regions of elevated failure probability as \emph{Positional Blind Spots} (PBS).
This observation raises two practical questions: how can PBS be uncovered under a black-box setting with limited rollout budgets, and how can the discovered regions guide targeted policy refinement? 
The next subsection formalizes this problem.


\subsection{Problem Definition}

\subsubsection{VLA task execution.}
Consider a manipulation task $\mathcal{T}$ with language instruction $\ell$, a task-relevant target object $x^{tar}$, and a task-irrelevant distractor object $x^{dis}$, together with a task-adapted VLA policy $\pi_\theta$.
The policy is expected to manipulate $x^{tar}$ according to $\ell$, while ignoring $x^{dis}$.
Let $\Omega\subset\mathbb{R}^{2}$ denote the feasible placement region of the task-irrelevant distractor.
For each position $z\in\Omega$, we initialize the environment by placing the distractor at $z$, while keeping the task objective, language instruction, task-relevant object configuration, robot initialization, camera configuration, and distractor attributes unchanged.
During execution, the policy receives visual observations $o_t$ and predicts actions conditioned on the instruction $\ell$:
\begin{equation}
    a_t \sim \pi_\theta(\cdot \mid o_t,\ell).
\end{equation}
The generated actions induce a trajectory $\tau(z)$.
We define the binary execution outcome as $Y_\theta(z)\in\{0,1\}$, where $Y_\theta(z)=1$ indicates task failure and $Y_\theta(z)=0$ indicates successful completion.
The position-conditioned failure probability is then defined as
\begin{equation}
p_\theta(z)=P(Y_\theta(z)=1\mid z).
\end{equation}
Thus, $p_\theta(z)$ captures the failure risk induced by placing the same distractor at position $z$, while controlling other task and scene factors. A high value of $p_\theta(z)$ indicates that the distractor placement at $z$ substantially increases the likelihood of task failure compared with other feasible placements.

\subsubsection{Positional Blind Spots (PBS).} 
Based on the position-conditioned failure probability $p_\theta(z)$ of a VLA policy, we define \emph{Positional Blind Spots} (PBS) as spatial regions where placing the same task-irrelevant distractor leads to substantially higher failure risk than the average failure risk over the workspace.
Let $\mu$ denote a reference distribution over feasible distractor positions, instantiated as the uniform distribution over $\Omega$. The workspace-level failure probability is defined as
\begin{equation}
    \bar{p}_\theta
    =
    \mathbb{E}_{z\sim\mu}\left[p_\theta(z)\right].
\end{equation}
Given a risk margin $\delta>0$, we can define the theoretical PBS set as
\begin{equation}
    \mathcal{B}_\theta(\delta)
    =
    \left\{
        z\in\Omega
        \mid
        p_\theta(z)\geq\bar{p}_\theta+\delta
    \right\}.
\end{equation}
Each spatially connected component of $\mathcal{B}_\theta(\delta)$ corresponds to a Positional Blind Spot (PBS), representing a localized workspace region where the task-irrelevant distractor induces substantially higher failure risk than the average risk over the workspace.
This definition characterizes relative excess failure risk and requires no access to policy gradients, confidence scores, attention maps, or other internal model signals.
In practice, $\mathcal{B}_\theta(\delta)$ is unknown and must be estimated through black-box rollouts under a limited evaluation budget.

\subsubsection{PBS uncovering and mitigation problem.}
Given that the true PBS region $\mathcal{B}_\theta(\delta)$ is unknown, the PBS uncovering problem aims to identify an estimated region $\widehat{\mathcal{B}}$ using only black-box rollout outcomes under a limited evaluation budget $Q$, such that $\widehat{\mathcal{B}}$ closely approximates $\mathcal{B}_\theta(\delta)$ through efficient allocation of rollouts to informative positions.
Based on the discovered region $\hat{\mathcal{B}}$, the PBS mitigation problem aims to collect successful demonstrations at positions within $\hat{\mathcal{B}}$ and adapt the VLA policy to reduce failure risks within PBS regions while preserving performance on non-PBS regions.
This formulation decomposes the PBS problem into two stages: identifying spatial regions with elevated failure risks and improving policy performance within those regions. The following methodology instantiates these two stages.

\section{Methodology}
\subsection{Overview}

\begin{figure*}[!t]
    \centering
    \includegraphics[
        width=\textwidth,
    ]{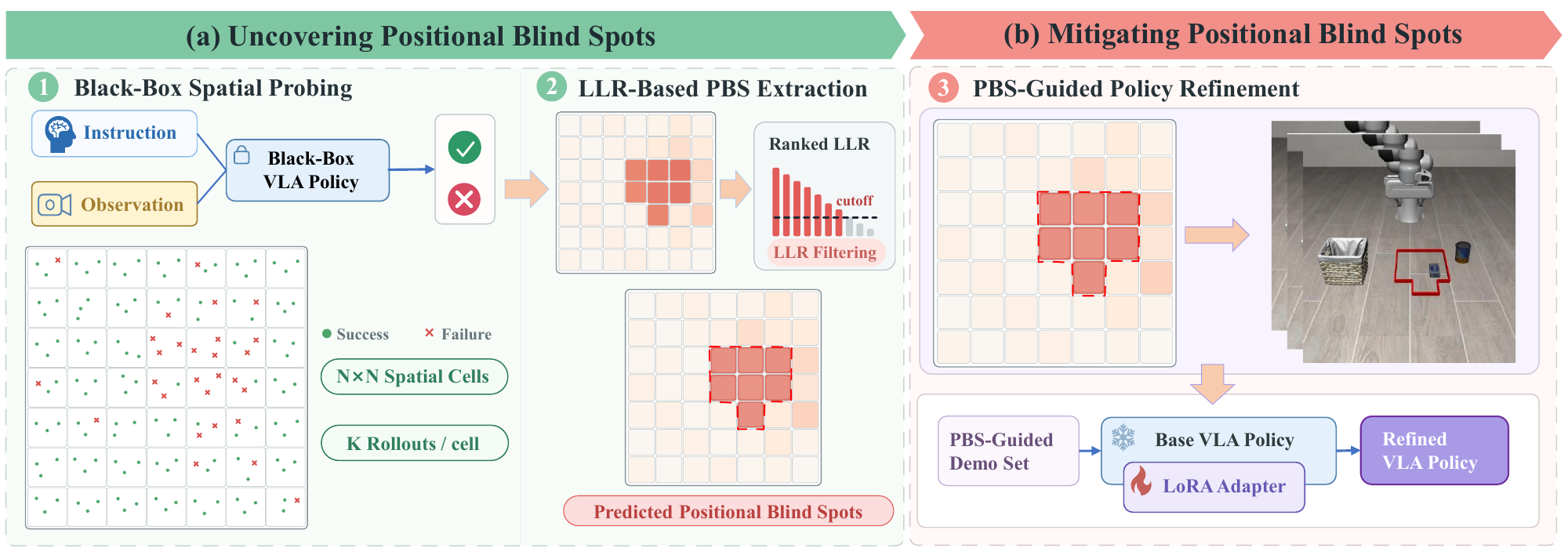}
    \caption{
    Pipeline of the proposed framework for uncovering and mitigating Positional Blind Spots (PBS).
    }
    \label{fig:overview}
\end{figure*}

Based on the problem definition, the distractor position is the only scene variable systematically changed, while the target object configuration, task objective, and VLA policy remain fixed. During evaluation, the policy is accessed solely through binary rollout outcomes.
As illustrated in Figure~\ref{fig:overview}, our framework consists of two major stages: PBS uncovering and PBS-guided mitigation. 
In the uncovering stage, we partition the feasible placement region into spatial cells and perform black-box rollouts by uniformly sampling positions within each cell. 
The collected rollout outcomes are then analyzed using an LLR-based extraction procedure to rank spatial cells according to their failure risk and identify the predicted PBS regions.
In the mitigation stage, we collect successful demonstrations from the discovered PBS regions and refine the VLA policy through LoRA-based adaptation, yielding a more robust policy against positional failures.


\subsection{Uncovering Positional Blind Spots}
\label{sec:uncovering}

The uncovering stage identifies spatial regions where the placement of the task-irrelevant distractor object $x^{dis}$ at position $z$ increases the failure risk of manipulating the target object $x^{tar}$.
We consider a black-box setting in which the policy is queried only through complete rollouts, and only binary success or failure outcomes are observed.
We formulate this process as a budgeted spatial search based on uniform sampling and statistical risk comparison.

Let $\Omega\subset\mathbb{R}^{2}$ denote the feasible distractor-placement region. 
We discretize $\Omega$ into an $N\times N$ uniform grid containing $N^2$ cells, denoted as $\{g_i\}_{i=1}^{N^2}$, such that
\begin{equation}
    \Omega=\bigcup_{i=1}^{N^2}g_i,\qquad
    g_i\cap g_j=\emptyset,\quad \forall i\neq j.
\end{equation}
For each cell $g_i$, we uniformly sample $K$ valid positions and execute one rollout at each position, resulting in a total budget of $Q=N^2K$ policy executions.
This discretization enables systematic coverage of the placement region while retaining sufficient spatial resolution to localize concentrated failures.
All task and scene factors, including the target object configuration, language instruction, and robot initialization, remain fixed except for the distractor position.
Let $c_i$ denote the number of failures observed in cell $g_i$, and let its empirical failure probability be $\hat{p}_i=c_i/K$, which provides a local estimate of failure risk for subsequent PBS identification.


The uniform allocation of rollouts across cells enables comparable estimation of failure risk across cells, as each cell is evaluated with the same number of trials.
However, empirical failure probability alone cannot distinguish a localized positional vulnerability from the policy's overall failure rate, since cells may exhibit high failure probabilities simply because the policy is generally difficult on the task.
We therefore assign each cell an LLR score to measure whether its observed failure rate represents a localized increase relative to the surrounding workspace.
For a cell $g_i$, the null hypothesis assumes that its failure probability is equal to that of the remaining workspace, whereas the alternative hypothesis assumes a higher failure probability within the cell.
The likelihoods are computed from the observed binomial failure counts, and the LLR score is set to zero when the empirical failure probability within the cell does not exceed that of the remaining workspace.
A higher LLR score indicates stronger evidence that the cell corresponds to a localized region with elevated failure risk, and the scores are subsequently used to rank cells for PBS extraction.


After computing the LLR score for each cell, we rank all cells in descending order and retain the $M$ highest-scoring cells, where $1\leq M\leq N^2$ denotes the number of selected cells.
Denoting the ordering by $\langle  g_{(1)},\ldots,g_{(N^2)}\rangle $, the predicted positional blind-spot set is
\begin{equation}
    \hat{\mathcal{B}}
    =
    \left\{
        g_{(1)},\ldots,g_{(M)}
    \right\}.
\end{equation}
The selection size $M$ determines the spatial support forwarded to the mitigation stage and is fixed across different policies under the same experimental setting.
Adjacent selected cells are merged into the same blind-spot region, whereas disconnected cell groups are treated as distinct blind spots.
The resulting region $\hat{\mathcal{B}}$ is used to guide targeted demonstration collection in the mitigation stage.


\subsection{Mitigating Positional Blind Spots}
\label{sec:mitigation}

After identifying PBS, we collect additional demonstrations from these high-risk areas and adapt the VLA policy to reduce positional failures while preserving its behavior over the remaining workspace.
The key idea is to augment the policy with successful executions under previously vulnerable distractor configurations, while keeping other task and scene factors unchanged.


Let $\hat{\mathcal{B}}$ denote the PBS region discovered by the uncovering stage.
We sample $N_{\mathrm{demo}}$ valid distractor positions $z_j\in\hat{\mathcal{B}}$ and place the distractor object $x^{dis}$ at each sampled position.
For each sampled position, we place the distractor object accordingly and teleoperate the robot to execute the target manipulation task, collecting a successful demonstration trajectory.
Let $\tau_j(z_j)=\big(o_t^{(j)},a_t^{(j)}\big)_{t=0}^{T_j}$ denote the collected demonstration trajectory.
The PBS-guided demonstration dataset is defined as
\begin{equation}
    \mathcal{D}_{\mathrm{PBS}}
    =
    \{\tau_j(z_j)\}_{j=1}^{N_{\mathrm{demo}}},
\end{equation}
where each collected trajectory satisfies $\chi(\tau_j(z_j))=1$.
The task objective, target object configuration, language instruction, and environment configuration remain unchanged, with the distractor position being the only factor different from the original fine-tuning data.
These demonstrations provide additional coverage of positional configurations where the policy previously exhibited elevated failure risks.


We then fine-tune the policy using the newly collected PBS-guided demonstrations, following its native adaptation procedure.
Let $\theta$ denote the frozen parameters of the original policy and $\phi$ denote the trainable LoRA parameters.
Using the policy's original training objective $\mathcal{L}_{\mathrm{native}}$, the LoRA parameters are optimized as
\begin{equation}
    \phi^{*}
    =
    \arg\min_{\phi}
    \frac{1}{N_{\mathrm{demo}}}
    \sum_{j=1}^{N_{\mathrm{demo}}}
    \mathcal{L}_{\mathrm{native}}
    \big(\pi_{\theta,\phi};\tau_j\big).
\end{equation}
Consequently, the mitigated policy is obtained as
$\pi_{\mathrm{mit}}=\pi_{\theta,\phi^{*}}$.
In particular, we retain the original observation space, action representation, optimization objective, and training hyperparameters, while updating only the parameters introduced by LoRA adaptation.


Overall, the proposed mitigation procedure first identifies PBS through black-box uncovering, then collects PBS-guided demonstrations, and finally adapts the VLA policy using the collected data.
This search-guided adaptation focuses the limited demonstration effort on vulnerable positional configurations while preserving the original task execution procedure.
By avoiding additional supervision over already reliable regions, the proposed strategy enables targeted policy refinement with limited demonstration resources.


\section{experiments}
\begin{table*}[t]
    \centering
    \small
    \renewcommand{\arraystretch}{1.16}
    \setlength{\tabcolsep}{3.8pt}
    \caption{
        Quantitative results of positional vulnerability before and after PBS-guided mitigation. 
    }
    \label{tab:characterization_mitigation}
    \begin{tabular*}{\textwidth}{@{\extracolsep{\fill}}l|*{6}{c}|*{6}{c}@{}}
        \toprule

        \multicolumn{1}{c|}{Dataset}
        & \multicolumn{6}{c|}{LIBERO}
        & \multicolumn{6}{c}{VLA-Arena} \\
        \cmidrule(r){1-1}
        \cmidrule(lr){2-7}
        \cmidrule(lr){8-13}

        \multicolumn{1}{c|}{Stage}
        & \multicolumn{3}{c}{Pre-Mitigation}
        & \multicolumn{3}{c|}{Post-Mitigation}
        & \multicolumn{3}{c}{Pre-Mitigation}
        & \multicolumn{3}{c}{Post-Mitigation} \\
        \cmidrule(r){1-1}
        \cmidrule(lr){2-4}
        \cmidrule(lr){5-7}
        \cmidrule(lr){8-10}
        \cmidrule(lr){11-13}

        Model / Metric
        & LLR $\downarrow$
        & Moran's $I$ $\downarrow$
        & FR $\downarrow$
        & LLR $\downarrow$
        & Moran's $I$ $\downarrow$
        & FR $\downarrow$
        & LLR $\downarrow$
        & Moran's $I$ $\downarrow$
        & FR $\downarrow$
        & LLR $\downarrow$
        & Moran's $I$ $\downarrow$
        & FR $\downarrow$ \\
        \midrule

        $\pi_0$
        & 644.03 & 0.70 & 0.27
        & \textbf{0.43} & \textbf{0.24} & \textbf{0.04}
        & 491.05 & 0.61 & 0.49
        & \textbf{7.32} & \textbf{0.29} & \textbf{0.16} \\

        $\pi_{0.5}$
        & 950.42 & 0.86 & 0.31
        & \textbf{8.17} & \textbf{0.28} & \textbf{0.16}
        & 55.51 & 0.21 & 0.36
        & \textbf{4.10} & 0.24 & \textbf{0.10} \\

        OpenVLA-OFT
        & 1954.03 & 0.75 & 0.33
        & \textbf{80.59} & \textbf{0.31} & \textbf{0.11}
        & 631.02 & 0.66 & 0.58
        & \textbf{5.95} & \textbf{0.27} & \textbf{0.26} \\

        UniVLA
        & 445.10 & 0.49 & 0.15
        & \textbf{1.25} & \textbf{0.29} & \textbf{0.09}
        & 969.56 & 0.45 & 0.24
        & \textbf{6.74} & \textbf{0.21} & \textbf{0.07} \\

        VLA-Adapter
        & 776.31 & 0.55 & 0.36
        & \textbf{8.58} & \textbf{0.22} & \textbf{0.15}
        & -- & -- & --
        & -- & -- & -- \\

        \bottomrule
    \end{tabular*}
    
\end{table*}

\subsection{Experiment Setup}

\paragraph{Models and Benchmarks.}
We evaluate five representative VLA models, including $\pi_0$, $\pi_{0.5}$, OpenVLA-OFT, UniVLA, and VLA-Adapter, on two manipulation benchmarks: LIBERO~\cite{liu2023libero} and VLA-Arena~\cite{zhang2026vlaarena}. On LIBERO, all five models are evaluated using their corresponding LIBERO-adapted checkpoints.
On VLA-Arena, we evaluate $\pi_0$, $\pi_{0.5}$, OpenVLA-OFT, and UniVLA using their corresponding VLA-Arena-adapted checkpoints. 
VLA-Adapter is not included in the VLA-Arena evaluation because a publicly available VLA-Arena-adapted checkpoint is unavailable.


\paragraph{Evaluation Metrics.}
We evaluate positional vulnerability using LLR, Global Moran's $I$, and failure rate (FR), which characterize regional risk contrast, spatial autocorrelation, and overall failure frequency, respectively.
PBS uncovering performance is evaluated using Precision, Recall, F1-score, and Intersection over Union (IoU).

To perform exhaustive spatial evaluation, we discretize the feasible distractor-placement workspace using the default distractor footprint as the cell size, yielding a $14\times14$ ground-truth evaluation grid $\mathcal{G}_{\mathrm{gt}}$.
For a candidate region $\mathcal{A}\subseteq\mathcal{G}_{\mathrm{gt}}$, let \((n_{\mathrm{in}},c_{\mathrm{in}})\) and \((n_{\mathrm{out}},c_{\mathrm{out}})\) denote the rollout and failure counts inside and outside $\mathcal{A}$.
The LLR score is computed as
\begin{equation}
    \mathrm{LLR}(\mathcal{A})
    =
    \mathbb{I}\!\left[
        \hat{p}_{\mathrm{in}}>\hat{p}_{\mathrm{out}}
    \right]
    \log
    \frac{\mathcal{L}(H_1;\mathcal{A})}
         {\mathcal{L}(H_0;\mathcal{A})},
\end{equation}
where \(\hat{p}_{\mathrm{in}}=c_{\mathrm{in}}/n_{\mathrm{in}}\) and \(\hat{p}_{\mathrm{out}}=c_{\mathrm{out}}/n_{\mathrm{out}}\).
We compute $\mathrm{LLR}_i=\mathrm{LLR}(\{g_i\})$ for each cell, rank positive scores in descending order, and apply Kneedle to identify the reference PBS region.

To further characterize the spatial structure of positional vulnerability, we compute Global Moran's $I$~\cite{burgert2025go} over cell-wise LLR scores:
\begin{equation}
    I
    =
    \frac{|\mathcal{G}|}{W}
    \frac{
        \sum_i\sum_j
        w_{ij}(x_i-\bar{x})(x_j-\bar{x})
    }{
        \sum_i(x_i-\bar{x})^2
    },
\end{equation}
where \(x_i=\mathrm{LLR}_i\), \(W=\sum_i\sum_j w_{ij}\), and \(w_{ij}\) denotes the four-neighbor spatial weight.
FR is computed as the fraction of failed rollouts over the exhaustive evaluation.
For PBS uncovering evaluation, the exhaustive evaluation-derived PBS region is treated as the reference, while $\hat{\mathcal{B}}$ is treated as the predicted region. Precision, Recall, F1-score, and IoU measure the localization accuracy and coverage.


\subsection{PBS Uncovering and Mitigation Performance}

\begin{figure*}[t]
    \centering
    \includegraphics[
        width=\textwidth,
    ]
    {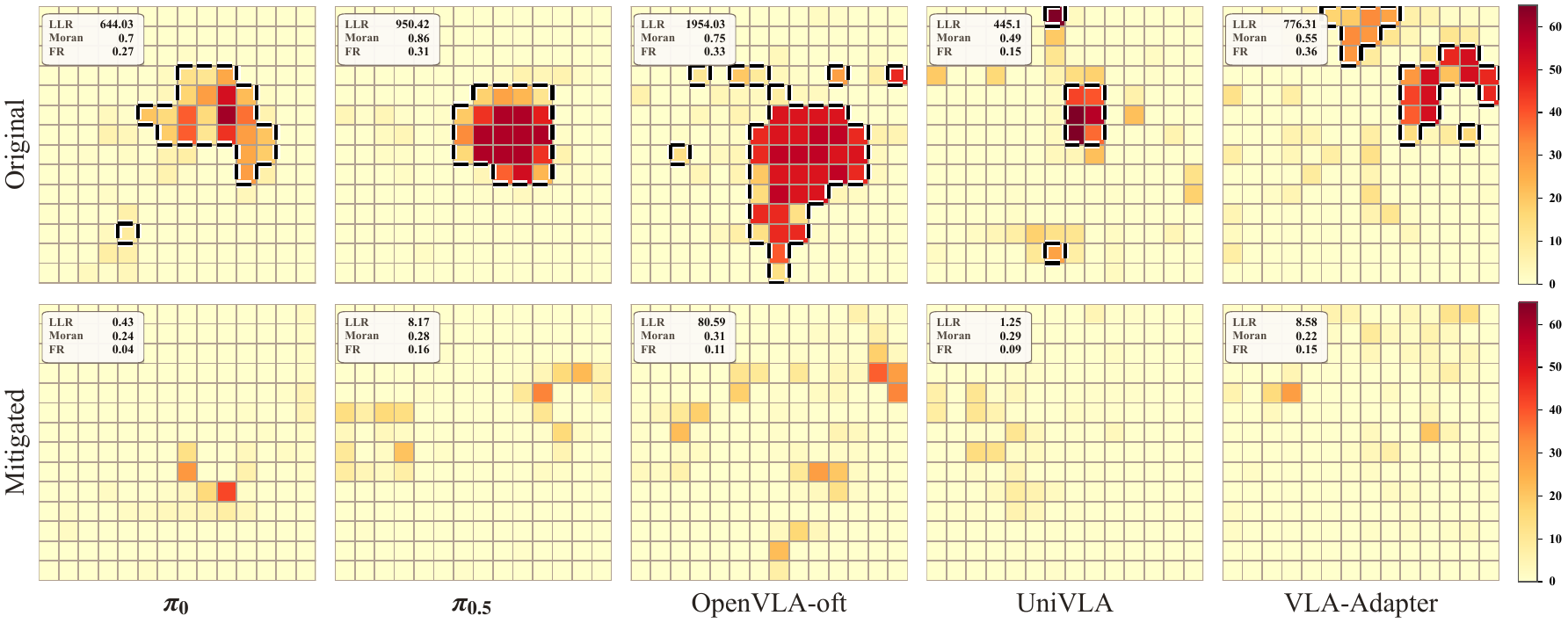}
    \caption{
    Visualization of cell-wise LLR distributions before and after PBS-guided mitigation. 
    }

    \label{fig:positional_llr_maps}
\end{figure*}

We first investigate whether task-irrelevant distractors induce systematic positional vulnerabilities in VLA policies.
As shown by the pre-mitigation results in Table~\ref{tab:characterization_mitigation}, all evaluated policies exhibit substantial positional failures, with FR ranging from $0.15$ to $0.58$ across LIBERO and VLA-Arena.
The high LLR values indicate that failures are not uniformly distributed across the workspace but concentrated in specific distractor-placement regions, while the positive Moran's $I$ values demonstrate clear spatial autocorrelation of these vulnerable regions.
Moreover, Figure~\ref{fig:positional_llr_maps} reveals that different VLA policies exhibit distinct high-risk locations and spatial extents, suggesting that PBS are architecture-dependent rather than solely determined by the task environment.


We next evaluate whether PBS-guided mitigation can effectively reduce positional vulnerability.
As shown by the post-mitigation results in Table~\ref{tab:characterization_mitigation}, PBS-guided fine-tuning consistently improves positional robustness across all evaluated settings.
Specifically, LLR decreases by $92.61\%$--$99.93\%$, indicating that the risk contrast between vulnerable and non-vulnerable regions is substantially reduced, while FR decreases by $40.00\%$--$85.19\%$, demonstrating improved overall task reliability.
Moran's $I$ also decreases for all policies on LIBERO and most policies on VLA-Arena, suggesting that the remaining failures become less spatially concentrated.
The only exception is $\pi_{0.5}$ on VLA-Arena, where Moran's $I$ slightly increases despite reductions in LLR and FR, indicating that a small number of residual failures remain locally correlated.
Consistently, Figure~\ref{fig:positional_llr_maps} shows that the previously concentrated high-LLR regions become substantially weaker after mitigation, confirming that PBS-guided demonstrations effectively target vulnerable positional configurations and improve positional robustness.


\subsection{Comparison with Existing Search Strategies}

In this section, we compare our method with two baseline sampling strategies under the same evaluation budget. 
Specifically, Full-Region Random Sampling draws uniformly over the workspace, while Two-Stage Adaptive Sampling runs five rollouts per cell, then assigns the rest to the top-10 cells. 
All methods operate on the same $7\times7$ search grid with a shared budget of 980 policy executions and return seven cells as the predicted PBS region.
As shown in Table~\ref{tab:search_performance}, our method consistently outperforms both baselines across all evaluated policies.
On average, our method achieves Precision, Recall, F1-score, and IoU of $0.636$, $0.824$, $0.678$, and $0.528$, respectively, improving F1-score by $0.268$ over Random Sampling and by $0.178$ over Adaptive Search.
The higher recall demonstrates that uniform cell-wise allocation provides more reliable risk estimation over the workspace and reduces the likelihood of missing localized high-risk regions.
In contrast, Random Sampling may suffer from insufficient observations for individual cells, while Adaptive Search may prematurely focus on regions with high initial but unreliable risk estimates.
By maintaining balanced exploration across candidate regions, our method achieves more accurate PBS localization under limited rollout budgets.
Overall, these results demonstrate that our approach effectively uncovers PBS using only 980 rollouts, requiring only one tenth of the executions used by exhaustive evaluation.


\subsection{Ablation and Generalization Analysis}


\begin{table}[t]
    \centering
    
    \small
    \renewcommand{\arraystretch}{1.05}
    \setlength{\tabcolsep}{2.8pt}
    \setlength{\arrayrulewidth}{0.4pt}
    \caption{
         PBS uncovering performance of different sampling strategies under the same rollout budget.
    }
    \label{tab:search_performance}
    \begin{tabularx}{\columnwidth}{@{}c|l|*{4}{>{\centering\arraybackslash}X}@{}}
        \toprule
        Model
        & Method
        & Prec. $\uparrow$
        & Rec. $\uparrow$
        & F1 $\uparrow$
        & IoU $\uparrow$ \\
        \midrule

        \multirow{3}{*}{$\pi_0$}
        & Random
        & 0.50 & 0.70 & 0.58 & 0.41 \\
        & Adaptive
        & 0.61 & 0.74 & 0.67 & 0.50 \\
        & Ours
        & \textbf{0.64}
        & \textbf{0.78}
        & \textbf{0.71}
        & \textbf{0.55} \\
        \cmidrule(lr){1-6}

        \multirow{3}{*}{$\pi_{0.5}$}
        & Random
        & 0.47 & 0.68 & 0.56 & 0.38 \\
        & Adaptive
        & 0.50 & 0.73 & 0.59 & 0.42 \\
        & Ours
        & \textbf{0.75}
        & \textbf{0.95}
        & \textbf{0.84}
        & \textbf{0.72} \\
        \cmidrule(lr){1-6}

        \multirow{3}{*}{OpenVLA-OFT}
        & Random
        & 0.43 & 0.35 & 0.39 & 0.24 \\
        & Adaptive
        & 0.71 & 0.59 & 0.65 & 0.48 \\
        & Ours
        & \textbf{0.93}
        & \textbf{0.63}
        & \textbf{0.75}
        & \textbf{0.60} \\
        \cmidrule(lr){1-6}

        \multirow{3}{*}{UniVLA}
        & Random
        & 0.07 & 0.25 & 0.11 & 0.06 \\
        & Adaptive
        & 0.14 & 0.50 & 0.22 & 0.13 \\
        & Ours
        & \textbf{0.29}
        & \textbf{1.00}
        & \textbf{0.44}
        & \textbf{0.29} \\
        \cmidrule(lr){1-6}

        \multirow{3}{*}{VLA-Adapter}
        & Random
        & 0.36 & 0.47 & 0.41 & 0.26 \\
        & Adaptive
        & 0.32 & 0.43 & 0.37 & 0.23 \\
        & Ours
        & \textbf{0.57}
        & \textbf{0.76}
        & \textbf{0.65}
        & \textbf{0.48} \\
        \midrule

        \multirow{3}{*}{Average}
        & Random
        & 0.366 & 0.490 & 0.410 & 0.270 \\
        & Adaptive
        & 0.456 & 0.598 & 0.500 & 0.352 \\
        & Ours
        & \textbf{0.636}
        & \textbf{0.824}
        & \textbf{0.678}
        & \textbf{0.528} \\
        \bottomrule
    \end{tabularx}
    
\end{table}

\begin{table}[t]
    \centering
    \caption{Effect of search configuration on PBS uncovering.}
    \label{tab:search_configuration}
    \small
    \renewcommand{\arraystretch}{1.0}
    \setlength{\tabcolsep}{5.0pt}
    \setlength{\arrayrulewidth}{0.4pt}
    \begin{tabularx}{\columnwidth}{@{}l|*{5}{>{\centering\arraybackslash}X}@{}}
        \toprule
        Metric / $N$
        & 4
        & 5
        & 6
        & 7
        & 8 \\
        \midrule
        Precision $\uparrow$
        & 0.38 & 0.48 & 0.58 & \textbf{0.75} & 0.55 \\
        Recall $\uparrow$
        & 0.84 & 0.86 & 0.86 & \textbf{0.95} & 0.62 \\
        F1-score $\uparrow$
        & 0.52 & 0.62 & 0.69 & \textbf{0.84} & 0.58 \\
        IoU $\uparrow$
        & 0.35 & 0.45 & 0.53 & \textbf{0.72} & 0.41 \\
        \bottomrule
    \end{tabularx}
    
\end{table}

\begin{table}[t]
    \centering
    \caption{Effect of mitigation rounds on positional robustness.}
    \label{tab:finetuning_rounds}
    \small
    \renewcommand{\arraystretch}{1.05}
    \setlength{\tabcolsep}{5.5pt}
    \setlength{\aboverulesep}{0.35ex}
    \setlength{\belowrulesep}{0.35ex}
    \setlength{\arrayrulewidth}{0.4pt}
    \begin{tabularx}{\columnwidth}{@{}l|*{4}{>{\centering\arraybackslash}X}@{}}
        \toprule
        Metric / $r$
        & 0
        & 1
        & 2
        & 3 \\
        \midrule
        LLR $\downarrow$
        & 950.42
        & 8.17
        & 3.47
        & \textbf{2.64} \\
        Moran's $I$ $\downarrow$
        & 0.86
        & 0.28
        & \textbf{0.21}
        & 0.22 \\
        FR $\downarrow$
        & 0.31
        & 0.16
        & \textbf{0.08}
        & 0.10 \\
        \bottomrule
    \end{tabularx}
    
\end{table}

\begin{table}[t]
    \centering
    \caption{Generalization of $\pi_{0.5}$ under different adaptations.}
    \label{tab:augmented_dataset_comparison}
    \small
    \renewcommand{\arraystretch}{1.05}
    \setlength{\tabcolsep}{1.5pt}
    \setlength{\arrayrulewidth}{0.4pt}

    \begin{tabular*}{\columnwidth}{
        @{\extracolsep{\fill}}
        l|cccc|c
        @{}
    }
        \toprule
        Metric
        & LIBERO
        & VLA-Arena
        & VLABench
        & Average
        & Ours \\
        \midrule

        LLR \(\downarrow\)
        & 950.42
        & 55.51
        & 539.34
        & 515.09
        & \textbf{8.17} \\

        Moran's \(I\) \(\downarrow\)
        & 0.86
        & \textbf{0.21}
        & 0.73
        & 0.60
        & 0.28 \\

        FR \(\downarrow\)
        & 0.31
        & 0.36
        & 0.33
        & 0.33
        & \textbf{0.16} \\

        \bottomrule
    \end{tabular*}
    
\end{table}


\begin{table}[t]
    \centering
    \small
    \caption{Cross-benchmark generalization of fine-tuned VLA.}
    \label{tab:cross_benchmark_generalization}
    \setlength{\arrayrulewidth}{0.4pt}
    \begin{tabularx}{\columnwidth}{@{}l|*{2}{>{\centering\arraybackslash}X}|*{2}{>{\centering\arraybackslash}X}@{}}
        \toprule
        Benchmark
        & \multicolumn{2}{c|}{VLABench}
        & \multicolumn{2}{c}{VLA-Arena} \\
        \cmidrule(r){1-1}
        \cmidrule(lr){2-3}
        \cmidrule(lr){4-5}
        Policy / Metric
        & SR $\uparrow$
        & IS $\uparrow$
        & SR $\uparrow$
        & IS $\uparrow$ \\
        \midrule
        Original
        & 0.00
        & 0.28
        & 0.04
        & 0.43 \\
        Fine-tuned
        & 0.00
        & \textbf{0.32}
        & \textbf{0.14}
        & \textbf{0.60} \\
        \bottomrule
    \end{tabularx}
    
\end{table}

\paragraph{Effect of search granularity.} We investigate the effect of search granularity on PBS uncovering using $\pi_{0.5}$ on LIBERO by varying the workspace discretization resolution.
Specifically, we divide the workspace into an $N\times N$ grid and select the highest-ranked cells according to one-sided LLR scores.
We test $N\in\{4,5,6,7,8\}$ with $K=20$ executions per cell, resulting in a budget of $Q=20N^2$.
As shown in Table~\ref{tab:search_configuration}, $N=7$ achieves the best overall localization performance across all metrics.
Coarser grids provide insufficient spatial resolution and may merge neighboring but distinct vulnerable regions, whereas finer grids require substantially more rollouts to estimate cell-wise risks without consistent performance gains.
Therefore, we adopt $N=7$ as the default configuration for PBS uncovering.


\paragraph{Effect of mitigation rounds.}
Table~\ref{tab:finetuning_rounds} summarizes the effect of mitigation rounds on positional robustness using $\pi_{0.5}$ on LIBERO.
The first mitigation round substantially reduces all metrics, decreasing LLR from $950.42$ to $8.17$, Moran's $I$ from $0.86$ to $0.28$, and FR from $0.31$ to $0.16$.
Further rounds continue to reduce LLR but provide limited and non-monotonic improvements in Moran's $I$ and FR.
These results show that most positional vulnerabilities can be corrected through a single targeted adaptation, and we therefore adopt one mitigation round as a practical trade-off between robustness and adaptation cost.


\paragraph{Generalization across adaptation strategies.}
Table~\ref{tab:augmented_dataset_comparison} shows the positional robustness of $\pi_{0.5}$ checkpoints adapted on different datasets.
The three existing adaptation strategies exhibit high positional vulnerability, with an average LLR of $515.09$, Moran's $I$ of $0.60$, and FR of $0.33$.
The PBS-guided checkpoint reduces LLR to $8.17$ and FR to $0.16$ while maintaining a low Moran's $I$ of $0.28$, showing the benefit of targeted positional supervision.
Although the VLA-Arena-adapted checkpoint achieves the lowest Moran's $I$ among existing adaptation checkpoints, its higher FR than the LIBERO-adapted checkpoint shows that lower spatial autocorrelation does not necessarily imply fewer failures.


\paragraph{Generalization across benchmarks.}
We evaluate the original and PBS-guided fine-tuned $\pi_{0.5}$ policies on unseen benchmarks using success rate (SR) and normalized soft instruction score (IS).
Table~\ref{tab:cross_benchmark_generalization} shows improved performance on VLA-Arena and higher VLABench IS after PBS-guided fine-tuning, indicating that the proposed mitigation can generalize beyond the original LIBERO benchmark.



\section{Conclusion}

In this work, we identify Positional Blind Spots (PBS), localized regions where relocating a task-irrelevant distractor systematically increases failure risk in VLA policies.
We propose a two-stage framework that uncovers PBS through spatial probing and LLR-based analysis, and mitigates them using PBS-guided demonstrations and LoRA adaptation.
Experiments on five VLA policies across two benchmarks demonstrate that our framework accurately localizes PBS regions and effectively reduces positional vulnerability.


\bibliographystyle{IEEEtran}
\bibliography{references}

\end{document}